\documentclass[conference]{IEEEtran}
\usepackage{ragged2e}
\IEEEoverridecommandlockouts
\usepackage{cite}
\usepackage{amsmath,amssymb,amsfonts}
\usepackage{algorithmic}
\usepackage[ruled,vlined]{algorithm2e}
\usepackage{graphicx}
\usepackage{textcomp}
\usepackage{subfigure}

\usepackage{fancyhdr}
\usepackage{amsmath}
\usepackage{floatrow} 
\usepackage[vlines]{tabularht}
\usepackage{pbox}
\usepackage{pifont}
\usepackage{multirow}
\usepackage[table,xcdraw]{xcolor}
\usepackage{floatrow}
\usepackage{adjustbox}
\usepackage{multicol}
\usepackage[hidelinks]{hyperref}
\usepackage{dirtytalk}
\usepackage{graphicx}
\usepackage{multirow}
\usepackage{lipsum}
\usepackage{makecell}
\usepackage{longtable}
\usepackage{tabularx}

\usepackage{array}
\usepackage{ragged2e}  

\usepackage{booktabs}
\usepackage{times}

\usepackage{mathptmx}
\usepackage{anyfontsize}

\def\BibTeX{{\rm B\kern-.05em{\sc i\kern-.025em b}\kern-.08em
    T\kern-.1667em\lower.7ex\hbox{E}\kern-.125emX}}

\makeatletter
\newcommand{\newlineauthors}{%
  \end{@IEEEauthorhalign}\hfill\mbox{}\par
  \mbox{}\hfill\begin{@IEEEauthorhalign}
}
\makeatother

\begin{document}

\title{SwinTextUNet: Integrating CLIP-Based Text Guidance into Swin Transformer U-Nets for Medical Image Segmentation\\
}

\author{
\IEEEauthorblockN{Ashfak Yeafi}
\IEEEauthorblockA{Department of Electrical and Electronic Engineering\\
Khulna University of Engineering \& Technology\\
Khulna-9203, Bangladesh\\
Email: yeafiashfak@gmail.com}
\and
\IEEEauthorblockN{Parthaw Goswami}
\IEEEauthorblockA{Department of Electronics and Communication Engineering\\
Khulna University of Engineering \& Technology\\
Khulna-9203, Bangladesh\\
Email: parthawgoswami555@gmail.com}
\newlineauthors
\IEEEauthorblockN{Md Khairul Islam}
\IEEEauthorblockA{Department of Mathematics and Computer Science\\
Hobart and William Smith Colleges\\
Geneva, NY, USA\\
Email: khairul.robotics@gmail.com}
\and
\IEEEauthorblockN{Ashifa~Islam~Shamme}
\IEEEauthorblockA{Department of Electrical and Electronic Engineering\\
Khulna University of Engineering \& Technology\\
Khulna-9203, Bangladesh\\
Email: ashifa54islam@gmail.com}
}

\IEEEaftertitletext{\vspace{-0.9cm}}
\maketitle

% Define a temporary header for the first page
\fancypagestyle{firstpagestyle}{
  \fancyhf{} % Clear header/footer
  \lhead{\fontsize{9}{11}\selectfont\justify 2025 28th International Conference on Computer and Information Technology (ICCIT)\\19-21 December 2025, Cox’s Bazar, Bangladesh} % Add conference name to the center of the header
  
    \fancyfoot[L]{979-8-3315-7867-1/25/\$31.00~\copyright2025 IEEE}
  \renewcommand{\headrulewidth}{0pt}
}
\thispagestyle{firstpagestyle}

\thispagestyle{firstpagestyle}

\begin{abstract}
Precise medical image segmentation is fundamental for enabling computer-aided diagnosis and effective treatment planning. Traditional models that rely solely on visual features often struggle when confronted with ambiguous or low-contrast patterns. To overcome these limitations, we introduce SwinTextUNet, a multimodal segmentation framework that incorporates Contrastive Language-Image Pre-training (CLIP), derived textual embeddings into a Swin Transformer U-Net backbone. By integrating cross-attention and convolutional fusion, the model effectively aligns semantic text guidance with hierarchical visual representations, enhancing robustness and accuracy. We evaluate our approach on the QaTa-COV19 dataset, where the proposed four-stage variant achieves an optimal balance between performance and complexity, yielding Dice and IoU scores of 86.47\% and 78.2\%, respectively. Ablation studies further validate the importance of text guidance and multimodal fusion. These findings underscore the promise of vision–language integration in advancing medical image segmentation and supporting clinically meaningful diagnostic tools.
\end{abstract}


\section{Introduction}
Precise segmentation of medical images is crucial for computer-assisted diagnosis (CAD), disease measurement, and treatment planning. Classical CNN-based models, such as U-Net and its extensions, have achieved strong results attributed to their encoder–decoder architecture incorporating skip connections. However, these frameworks rely solely on visual features, ignoring textual information (e.g., radiology reports or diagnostic notes) that clinicians routinely use to interpret images. This limitation reduces robustness when image features are ambiguous or noisy.  
Recent advances in transformer architectures, particularly the Swin Transformer~\cite{liu2021swin}, have enabled long-range dependency modeling with hierarchical efficiency, outperforming CNNs in both natural and medical image segmentation. Parallelly, vision–language models (VLMs) such as CLIP~\cite{radford2021learning} have demonstrated remarkable cross-modal alignment by jointly training image and text encoders. Despite their success in natural image domains, CLIP’s integration into medical segmentation remains underexplored.
In order to fill this void, we suggest SwinTextUNet, a multimodal segmentation framework that embeds CLIP-derived textual features into a Swin Transformer U-Net backbone. By fusing semantic text cues with visual features, our model improves segmentation robustness and captures clinically meaningful context.  
The key contributions of this work can be summarized as follows:
\begin{enumerate}
    \item A Swin Transformer U-Net variant that integrates CLIP-based textual embeddings for multimodal medical segmentation.  
    \item A text-conditioning mechanism that fuses semantic embeddings with image tokens across encoder and decoder stages for enhanced cross-modal alignment.  
\end{enumerate}
This paper's is structured as follows:  The work is reviewed in Section~\ref{sec:related_work}, the methodology is described in Section~\ref{sec:methodology}, the experimental results are reported in Section~\ref{sec:results}, and the work is concluded in Section~\ref{conl}.

\section{Related work}
\label{sec:related_work}
Medical image segmentation has evolved through multiple architectural paradigms, ranging from convolutional encoder–decoders to transformer-based models and, more recently, multimodal vision–language approaches. In this section, we review these developments, with a particular focus on biomedical segmentation networks that are closely related to our proposed framework.
\subsection{CNN-Based Architectures}
The primary model for medical segmentation is still the U-Net~\cite{ronneberger2015u}, which uses an encoder–decoder with skip links to maintain spatial detail. Numerous variants have extended this baseline: UNet++~\cite{zhou2018unet++} redesigned skip pathways for improved semantic fusion.
Specialized CNN extensions have been proposed for specific clinical domains. For example, ADTNet~\cite{yeafi2024adtnet} introduced an Attention-Guided U-Net with Dynamic Convolution and Transformers for skin cancer segmentation. Similarly, GSNet~\cite{jawad2023gsnet} and DSNet~\cite{yeafi2025deep} leveraged 3D convolutions and attention-based skip connections for glioma segmentation in the BraTS benchmarks. These works illustrate the potential of augmenting U-Net structures with domain-specific modules for improved performance in challenging segmentation tasks~\cite{parthaw1,parthaw2,parthaw3}.
\subsection{Transformer-Based Architectures}
Transformers~\cite{vaswani2017attention} have provided a powerful alternative to CNNs by capturing global context through self-attention. Their adaptation to medical segmentation has been highly impactful. TransUNet~\cite{chen2021transunet} combined CNN backbones with ViT encoders to balance local and global feature extraction. Swin-UNet~\cite{cao2022swin}, based on hierarchical Swin Transformers~\cite{liu2021swin}, introduced shifted window attention for scalable context modeling across resolutions. Further variants such as MISSFormer~\cite{huang2021missformer}, MedT~\cite{qi2022medt} refined hierarchical attention, token aggregation, and gated mechanisms to enhance boundary delineation. Model like EF-SwinNet~\cite{sarker2024ef} uses a hybrid architecture combining CNN and transformer for medical image analysis. While these architectures significantly outperform only CNN-based models, they are still limited to unimodal (image-only) input.
\subsection{Vision–Language Segmentation}
The emergence of large-scale contrastive pretraining has sparked interest in multimodal segmentation~\cite{khairul1}. CLIP~\cite{radford2021learning} demonstrated powerful alignment between vision and text embeddings. Medical adaptations such as BioViL~\cite{bannur2023learning}, PubMedCLIP~\cite{eslami2023pubmedclip} extended these models to radiology data, enabling classification, retrieval, and report understanding. For segmentation, recent works integrated CLIP with pixel-level architectures, MedCLIP-SAM~\cite{koleilat2024medclip} fused CLIP features with the Segment Anything Model for general-purpose medical segmentation. Although promising, these models often rely on external foundation modules or operate in weakly supervised settings\cite{yeafi2023semi}, limiting their applicability in specialized domains.
In contrast, our proposed SwinTextUNet introduces a fully supervised, end-to-end multimodal segmentation framework. By fusing CLIP-based text embeddings with hierarchical Swin Transformer U-Net features through cross-attention and ConvFuse blocks, our approach explicitly aligns textual priors with multiscale visual features. Unlike prior CNN-based or transformer-based methods, SwinTextUNet leverages both modalities to enhance segmentation quality, representing a novel contribution in medical vision–language modeling.
\section{Methodology}
\label{sec:methodology}
\begin{figure}[t]
    \centering
    \includegraphics[width=0.7\linewidth]{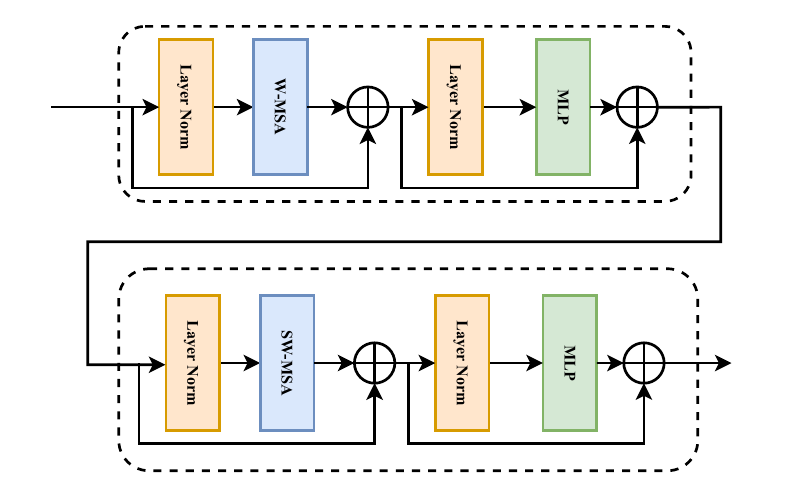}
    \caption{Illustration of the Swin Transformer block with windowed and shifted window attention.}
    \label{fig:swinblock}
\end{figure}

\begin{figure}[t]
    \centering
    \includegraphics[width=.85\linewidth]{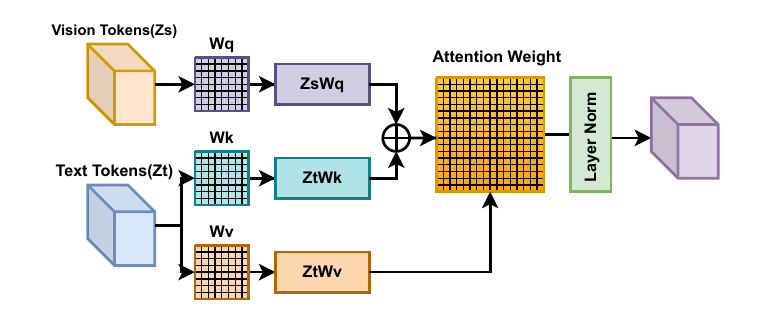}
    \caption{Cross-Attention block. Vision tokens attend to text tokens, refining representations with semantic guidance.}
    \label{fig:crossattention}
\end{figure}

\begin{figure}[t]
    \centering
    \includegraphics[width=0.8\linewidth]{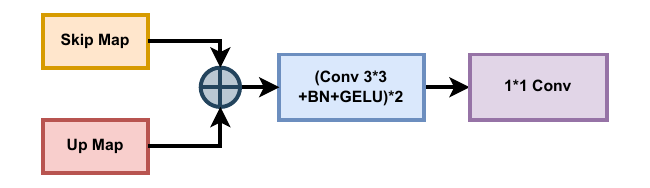}
    \caption{ConvFuse block. Skip map and upsampled tokens are concatenated, refined by convolutions, and re-tokenized.}
    \label{fig:convfuse}
\end{figure}

\begin{figure*}[!htbp]
  \centering
  \includegraphics[width=0.65\textwidth]{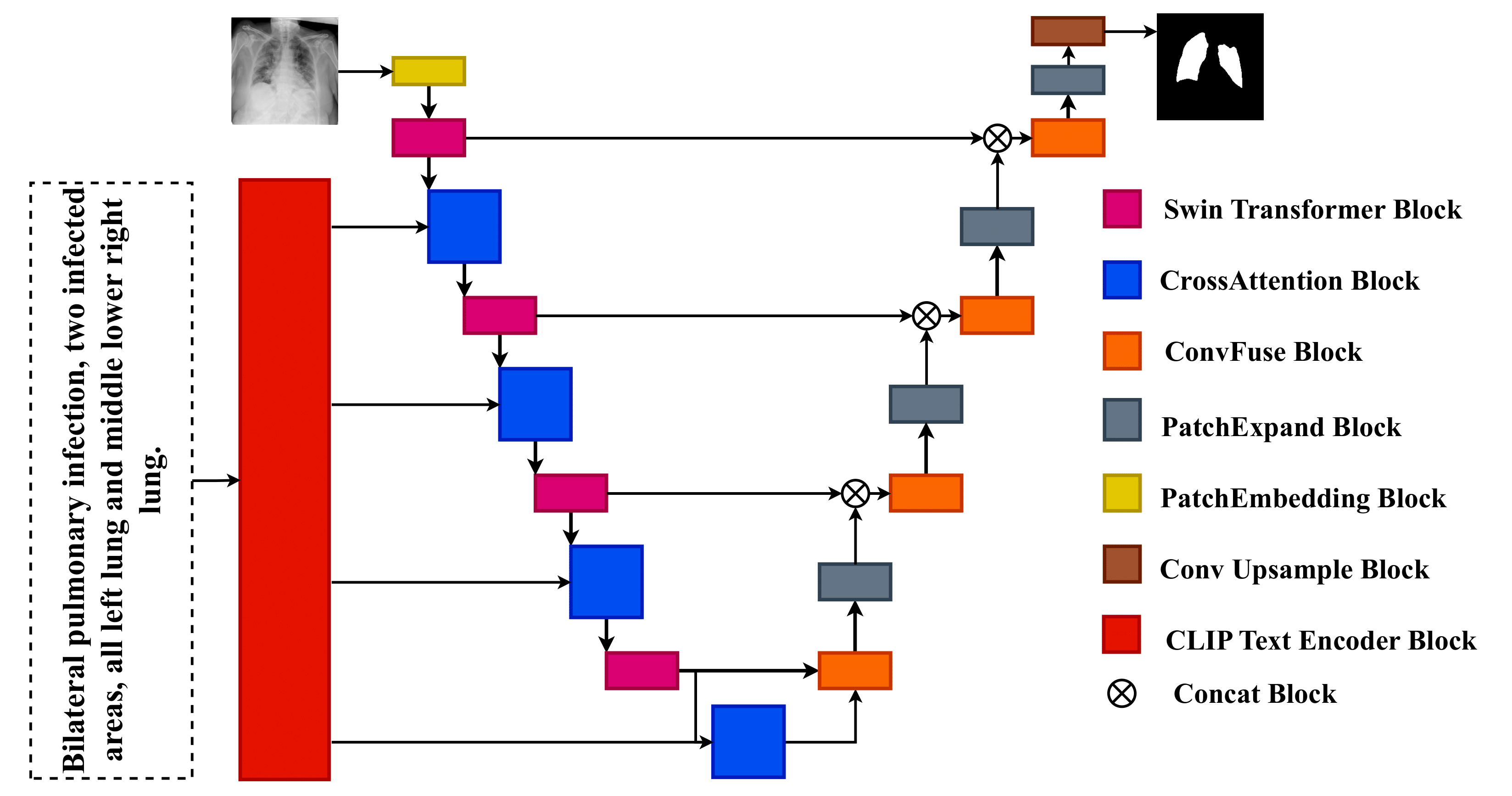}
  \caption{Overall architecture of SwinTextUNet. The model integrates CLIP-based text embeddings with a hierarchical Swin U-Net through cross-attention and ConvFuse blocks.}
  \label{fig:architecture}
\end{figure*}

We propose SwinTextUNet, a multimodal segmentation framework that integrates CLIP-based textual guidance into a hierarchical Swin Transformer U-Net. The model is designed to capture multi-scale visual features while incorporating semantic priors from domain-specific medical text. Four primary components make up the architecture, as seen in Figure~\ref{fig:architecture}: (i) CLIP-based text encoding, (ii) a hierarchical Swin Transformer encoder, (iii) text-guided cross-attention and convolutional fusion modules, and (iv) a decoder that reconstructs full-resolution segmentation masks. Key building blocks are further highlighted in Figures~\ref{fig:swinblock} (Swin Transformer Block), \ref{fig:crossattention} (Cross-Attention Block), and \ref{fig:convfuse} (ConvFuse Block).
\subsection{Text Encoding via CLIP}
To incorporate semantic guidance, we employ a pretrained CLIP text encoder~\cite{radford2021learning}. Given a batch of $B$ text prompts 
\[
\mathcal{T}=\{t^{(b)}\}_{b=1}^B,
\]
the encoder outputs pooled embeddings $\bar{Z}_t\in\mathbb{R}^{B\times D_t}$. Since OpenAI CLIP returns a single pooled vector, we form a token sequence by unsqueezing it, i.e., $Z_t\in\mathbb{R}^{B\times 1\times D_t}$. A linear projection aligns dimensions:
\begin{equation}
    \tilde{Z}_t = Z_t W_t, \quad \tilde{\bar{Z}}_t = \bar{Z}_t W_t, \quad W_t\in\mathbb{R}^{D_t\times D_v},
\end{equation}
where $D_v$ is the visual token dimension. The text tower remains frozen to preserve semantic priors, while $W_t$ is optimized during training.

\subsection{Visual Encoding with Swin Transformer}
The visual encoder is built upon the hierarchical Swin Transformer~\cite{liu2021swin}. The input image $X \in \mathbb{R}^{B \times 3 \times H \times W}$ is first divided into non-overlapping grous of shape $P \times P$ and projected into tokens:
\begin{equation}
    Z_1 \in \mathbb{R}^{B \times N_1 \times C_1}, \quad 
    N_1=\tfrac{H}{P}\cdot\tfrac{W}{P}, \; C_1=96, \quad (P=4).
\end{equation}
At each subsequent stage, the spatial resolution is halved while the channel dimension is doubled:
\begin{equation}
\begin{aligned}
Z_1 &\in \mathbb{R}^{B \times (56^2) \times 96}, \\
Z_2 &\in \mathbb{R}^{B \times (28^2) \times 192}, \\
Z_3 &\in \mathbb{R}^{B \times (14^2) \times 384}, \\
Z_4 &\in \mathbb{R}^{B \times (7^2) \times 768},
\end{aligned}
\end{equation}
for input resolution $224 \times 224$. The model can capture both global semantic context and fine-grained information attributable to its hierarchical nature. For decoding, stage outputs are reshaped into feature maps $S_s \in \mathbb{R}^{B \times C_s \times H_s \times W_s}$, which are later used as skip connections.

\subsection{Shifted-Window Self-Attention}
Each Swin block applies self-attention within local $M{\times}M$ windows:
\begin{equation}
    \mathrm{Attn}(Q,K,V) = \mathrm{Softmax}\!\left(\tfrac{QK^\top}{\sqrt{d_k}}+B_{\mathrm{rel}}\right)V,
\end{equation}
where $B_{\mathrm{rel}}$ is the relative position bias. A cyclic shift of $(M/2,M/2)$ between consecutive blocks enables inter-window connections, while a precomputed attention mask prevents invalid interactions. This reduces computational complexity from global $\mathcal{O}((HW)^2)$ to local $\mathcal{O}(HW\cdot M^2)$, enabling efficient high-resolution processing.
\subsection{Text-Guided Cross-Attention}
To integrate semantics, we refine vision tokens with text tokens at each stage $s$. Let $Z_s\in\mathbb{R}^{B\times N_s\times C_s}$ and $\tilde{Z}_t\in\mathbb{R}^{B\times T\times C_s}$ with $T=1$. Multi-head cross-attention computes:
\begin{align}
Q &= Z_s W_Q,\quad K = \tilde{Z}_t W_K,\quad V = \tilde{Z}_t W_V,\\
\tilde{Z}_s &= \mathrm{Softmax}\!\left(\tfrac{QK^\top}{\sqrt{d_k}}\right)V,\\
Z_s' &= \mathrm{LN}\!\big(Z_s + \tilde{Z}_s\big),\quad 
Z_s'' = Z_s' + \mathrm{MLP}(\mathrm{LN}(Z_s')).
\end{align}
This asymmetric design ensures that textual context steers visual features without overwhelming local structures. Complexity is $\mathcal{O}(BN_sTC_s)$, negligible since $T{\ll}N_s$.
\subsection{Convolutional Fusion (ConvFuse)}
The ConvFuse block merges upsampled decoder features $U_s$ with skip maps $S_s$ via convolutional refinement:
\begin{equation}
    F_s = \phi\!\big(\,[S_s,\,U_s]\,\big),
\end{equation}
where $[\,\cdot\,]$ denotes channel concatenation and $\phi$ consists of two $3{\times}3$ convolutions. Unlike pure token fusion, ConvFuse exploits local spatial continuity and improves boundary reconstruction. At the bottleneck, we additionally fuse $S_4$ into $Z_4$ to enrich deep features with fine-grained context.
\subsection{Decoder and Reconstruction}
Decoding proceeds via progressive PatchExpand followed by ConvFuse:
\[
\hat{Z}_4 \to Y_3 \to Y_2 \to Y_1 \to Y_0,
\]
with resolutions $\{7^2,14^2,28^2,56^2,112^2\}$. Tokens are converted to maps at each step, concatenated with skips, fused, and re-tokenized. The final map $Y_0\in\mathbb{R}^{B\times C_0/2\times 112\times112}$ is upsampled to $224\times224$ and processed through a $1{\times}1$ convolution to produce output.

\subsection{Output and Loss Function}
For binary segmentation, the output applies a sigmoid:
\begin{equation}
    \hat{Y} = \sigma(\hat{Y}_{\text{logits}}), \quad \sigma(x)=\frac{1}{1+e^{-x}}.
\end{equation}
We optimize a hybrid loss that combines cross-entropy and dice:
\begin{equation}
    \mathcal{L} = \lambda_{\text{dice}}\mathcal{L}_{\text{Dice}} + \lambda_{\text{ce}}\mathcal{L}_{\text{CE}},
\end{equation}
with
\begin{equation}
    \mathcal{L}_{\text{CE}} = -\tfrac{1}{N}\sum_{i=1}^N \big[y_i\log \hat{y}_i + (1-y_i)\log(1-\hat{y}_i)\big],
\end{equation}
\begin{equation}
    \mathcal{L}_{\text{Dice}} = 1-\frac{2\sum_i y_i\hat{y}_i+\epsilon}{\sum_i y_i+\sum_i\hat{y}_i+\epsilon}, \;\; \epsilon=10^{-6}.
\end{equation}
This balances overlap with per-pixel classification, alleviating class imbalance in medical segmentation.

\subsection{Training Pipeline}
Algorithm~\ref{alg:training} summarizes the complete training pipeline of SwinTextUNet.

\begin{algorithm}[h]
\caption{Training pipeline of SwinTextUNet}
\label{alg:training}
\KwIn{Image $X \in \mathbb{R}^{B \times 3 \times H \times W}$, text prompts $\mathcal{T}$, ground-truth mask $Y$.}
\KwOut{Predicted segmentation $\hat{Y}$.}
\BlankLine

\textbf{Step 1: Text Encoding}\\
Encode $\mathcal{T}$ with CLIP: $(Z_t, \bar{Z}_t) = f_{\text{text}}(\mathcal{T})$. \\
Project to visual dimension: $\tilde{Z}_t = W_t Z_t$. \\

\textbf{Step 2: Image Encoding}\\
Patch embed image $X$: $Z_0 = f_{\text{patch}}(X)$. \\
Run Swin Transformer blocks at each stage $s=1..4$ to obtain $Z_s$. \\
Extract skip maps $S_s$ for decoder. \\

\textbf{Step 3: Cross-Attention Fusion}\\
For each stage $s$, update tokens via $\hat{Z}_s = \text{CrossAttn}(Z_s,\tilde{Z}_t)$. \\

\textbf{Step 4: Decoder Reconstruction}\\
Initialize $Y_4=\hat{Z}_4$. For $s=4 \to 1$: \\
\quad Upsample $Y_s \to Y_{s-1}$. \\ 
\quad Fuse with skip $S_{s-1}$ using ConvFuse. \\

\textbf{Step 5: Output Layer}\\
Reconstruct full-resolution $Y_0$. \\ 
Compute logits: $\hat{Y} = \text{Conv}_{1\times1}(Y_0)$, apply sigmoid. \\

\textbf{Step 6: Loss Optimization}\\
Compute $\mathcal{L} = \lambda_{\text{dice}} \mathcal{L}_{\text{Dice}} + \lambda_{\text{ce}} \mathcal{L}_{\text{CE}}$. \\ 
Update weights with AdamW optimizer. \\

\end{algorithm}
\section{EXPERIMENT AND RESULT}
\label{sec:results}
\subsection{Dataset Description}
\label{sec:dataset}
\begin{figure}[h]
\centering
\includegraphics[width=.9\linewidth]{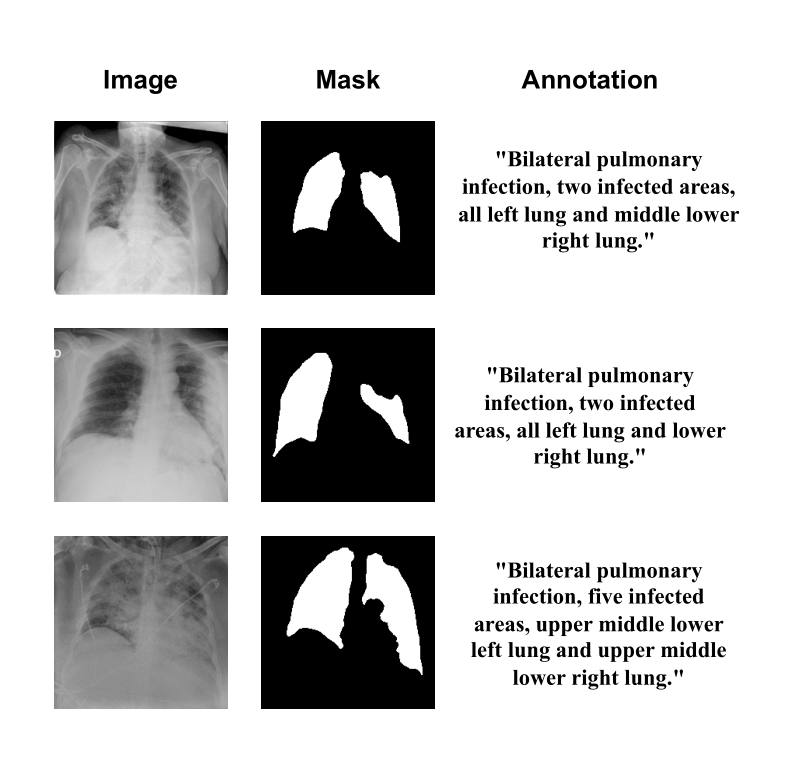}
\caption{Representative samples from the QaTa-COV19 dataset across training, validation, and test sets. Each sample shows (a) chest X-ray, (b) ground-truth lesion mask, and (c) textual annotation provided by the dataset authors~\cite{li2023lvit}.}
\label{fig:dataset_samples}
\end{figure}

We evaluate the proposed SwinTextUNet on the QaTa-COV19 dataset~\cite{degerli2022osegnet}, a large-scale benchmark of 9,258 chest X-ray (CXR) radiographs with manually annotated COVID-19 lesions. Each image is paired with a binary lesion mask delineating infected lung regions, enabling robust segmentation evaluation.
In addition, the dataset includes textual annotations curated by medical experts~\cite{li2023lvit}, describing infection patterns such as lesion count, anatomical location, and laterality (unilateral vs. bilateral). For example, \textit{``Bilateral pulmonary infection, two infected areas, upper left lung and upper right lung''}. These annotations are used as input to the CLIP text encoder, providing semantic priors to guide segmentation. Figure~\ref{fig:dataset_samples} illustrates sample triplets of CXR, mask, and annotation across subsets.
\subsection{Experimental Setup}
We evaluated the proposed SwinTextUNet on the QaTa-COV19 dataset~\cite{li2023lvit} using the official split of 5,716 training, 1,429 validation, and 2,113 test images. All chest X-rays and lesion masks were resized to $224 \times 224$, intensity-normalized to $[0,1]$, and masks were binarized. Text annotations were standardized into concise diagnostic prompts and processed through the CLIP tokenizer. The model was implemented in PyTorch and trained end-to-end for 100 epochs using the AdamW optimizer (learning rate $1 \times 10^{-4}$, weight decay $1 \times 10^{-2}$), cosine annealing with warmup, and a batch size of 8. A hybrid loss that combined dice and cross-entropy was used in the optimization process.  Random flips, rotations, and intensity scaling were used for data augmentation.
\subsection{Evaluation Metrics}
\label{eval}
We evaluate segmentation performance using the Dice Similarity Coefficient (Dice) and Intersection over Union (IoU), which quantify spatial overlap between predicted masks and ground truth. They are defined as:
\begin{equation}
    Dice = \frac{2TP}{2TP + FP + FN}, \qquad
    IoU = \frac{TP}{TP + FP + FN},
\end{equation}
where \(TP\), \(FP\), and \(FN\) denote true positives, false positives, and false negatives. Dice emphasizes region overlap, while IoU captures pixel-level agreement, making them well-suited for medical image segmentation.
\subsection{Result analysis}
% \label{sec:results}
This section provides a detailed evaluation of SwinTextUNet on the QaTa-COV19 dataset. The analysis covers segmentation accuracy, convergence behavior, architectural depth, ablation experiments, and comparisons with established baselines.
\subsubsection{Stage-Depth Exploration}
To investigate the effect of architectural depth, we experimented with 3-stage, 4-stage, and 5-stage variants of SwinTextUNet. The results, summarized in Table~\ref{tab:stage_depth}, indicate that the 3-stage variant provides acceptable accuracy but fails to capture fine-grained structures, leading to reduced Dice and IoU scores. The 5-stage variant slightly improves accuracy but introduces considerable parameter overhead, resulting in diminishing returns. By contrast, the 4-stage architecture achieves the best trade-off, with a Dice score of 86.47\% and IoU of 78.2\%, while maintaining a balanced parameter count. This confirms that a four-stage encoder–decoder structure is the most suitable configuration for the segmentation task.
\begin{table}[h]
\centering
\caption{Performance of SwinTextUNet with different stage depths.}
\label{tab:stage_depth}
\begin{tabular}{lccc}
\hline
\textbf{Variant} & \textbf{Dice (\%)} & \textbf{IoU (\%)} & \textbf{Params (M)} \\
\hline
3-Stage & 75.4 & 68.0 & 32.8 \\
\textbf{4-Stage (Ours)} & \textbf{86.47} & \textbf{78.2} & 57.6 \\
5-Stage & 87.1 & 78.8 & 105.4 \\
\hline
\end{tabular}
\end{table}
\subsubsection{Training Convergence}
\begin{figure}[h]
\centering
\includegraphics[width=.85\linewidth]{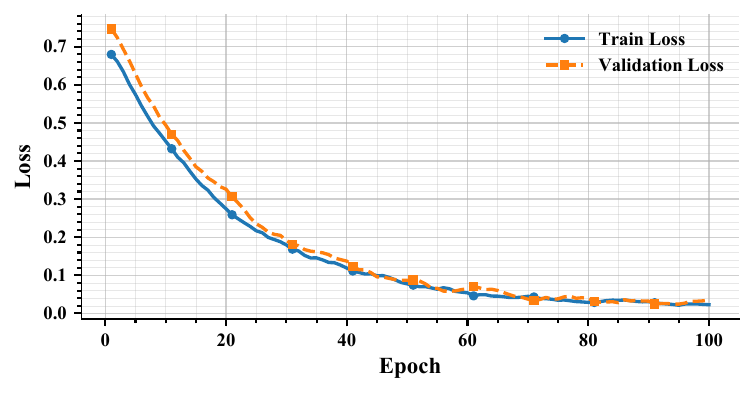}
\caption{Training and validation loss curves of SwinTextUNet.}
\label{fig:loss_curve}
\end{figure}

The convergence behavior of the 4-stage SwinTextUNet is illustrated in Figure~\ref{fig:loss_curve}. Both training and validation losses exhibit a smooth and monotonic decrease, with the validation loss closely following the training loss, suggesting minimal overfitting. Compared with 3-stage and 5-stage configurations, the 4-stage model demonstrates more stable convergence, which is consistent with its superior quantitative performance.
\subsubsection{Qualitative Results}
\begin{figure}[t]
\centering
\includegraphics[width=0.7\linewidth]{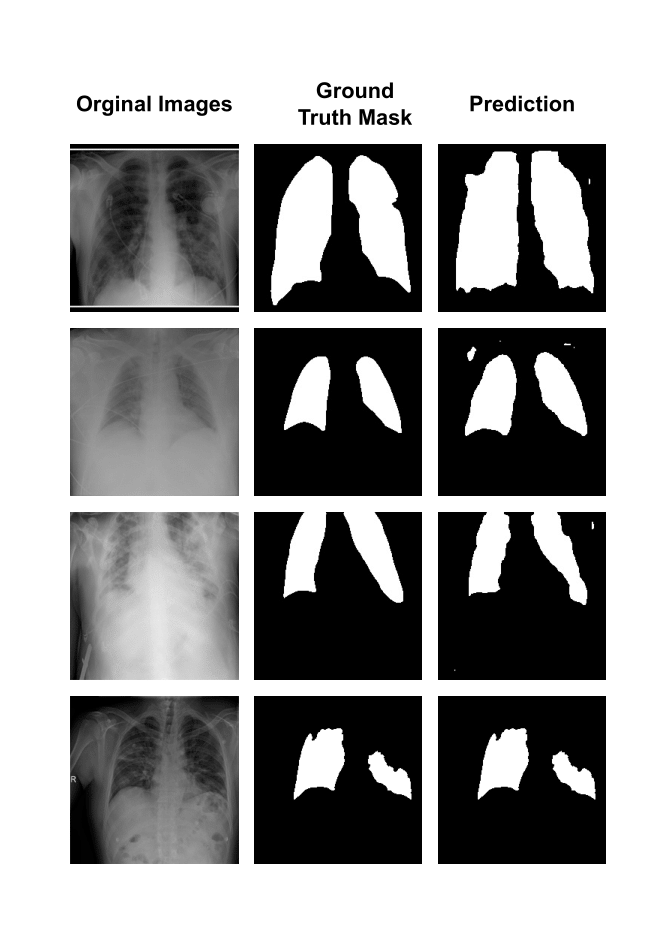}
\caption{Representative qualitative segmentation results. Each triplet shows (a) original CXR, (b) ground-truth mask, and (c) SwinTextUNet prediction.}
\label{fig:qualitative_results}
\end{figure}

Representative qualitative results are presented in Figure~\ref{fig:qualitative_results}. Each triplet shows an original chest X-ray, its ground-truth lesion mask, and the corresponding prediction from SwinTextUNet. The model accurately captures lesion boundaries, including complex bilateral and multi-focal infections. Predictions remain robust across varying lesion morphologies and sizes. Compared with standard Swin-UNet outputs, SwinTextUNet produces sharper boundaries and fewer false positives, highlighting the value of text-guided cross-attention in enhancing spatial alignment.
\subsubsection{Ablation Study}
To examine the contribution of individual modules, we performed an ablation study by selectively disabling text guidance, cross-attention, and ConvFuse. Results in Table~\ref{tab:ablation} show that removing text guidance leads to a substantial Dice drop of 7.3\% and an IoU drop of 8.8\%, underscoring the critical importance of semantic priors. Excluding ConvFuse decreases Dice by 3.1\% and IoU by 7.0\%, highlighting its role in effective multi-scale feature integration. Replacing cross-attention with simple concatenation causes a 1.8\% Dice and 2.8\% IoU reduction, demonstrating the necessity of explicit token-level alignment between modalities. The full SwinTextUNet consistently achieves the best results, validating the synergistic effect of all modules.
\begin{table}[h]
\centering
\caption{Ablation study of SwinTextUNet components.}
\label{tab:ablation}
\begin{tabular}{lcc}
\hline
\textbf{Model Variant} & \textbf{Dice (\%)} & \textbf{IoU (\%)} \\
\hline
w/o Text Guidance & 79.2 & 69.4 \\
w/o ConvFuse & 83.4 & 71.2 \\
w/o Cross-Attention & 84.7 & 75.4 \\
\textbf{Full SwinTextUNet} & \textbf{86.47} & \textbf{78.2} \\
\hline
\end{tabular}
\end{table}
\subsubsection{Comparison with Baseline Models}
Finally, SwinTextUNet was compared against widely used segmentation models. As shown in Table~\ref{tab:quantitative_results}, traditional CNN-based approaches underperform due to limited contextual modeling. Transformer-based methods achieve stronger results, yet SwinTextUNet surpasses all baselines, reaching 86.47\% Dice and 78.2\% IoU.
\begin{table}[h]
\centering
\caption{This section provides a detailed evaluation of SwinTextUNet on the QaTa-COV19 dataset.}
\label{tab:quantitative_results}
\begin{tabular}{lcc}
\hline
\textbf{Model} & \textbf{Dice (\%)} & \textbf{IoU (\%)} \\
\hline
U-Net~\cite{ronneberger2015u} & 79.02 & 69.46 \\
Attention U-Net~\cite{oktay2018attention} & 79.62 & 70.25 \\
CLIP~\cite{radford2021learning} & 79.81 & 70.66 \\
LViT~\cite{li2023lvit} & 83.66 & 75.71 \\
\textbf{SwinTextUNet (Ours)} & \textbf{86.47} & \textbf{78.2} \\
\hline
\end{tabular}
\end{table}
\subsection{Discussion}
Overall, SwinTextUNet demonstrates consistent improvements over both CNN- and Transformer-based baselines. The 4-stage configuration emerges as the most effective depth, balancing performance and complexity. Ablation experiments emphasize the necessity of multimodal fusion components, while loss curve analysis confirms stable training dynamics. Overall, the results confirm the effectiveness of SwinTextUNet and its promise for clinical use where precision and interpretability are crucial.

\section{Conclusion}
\label{conl}
We proposed SwinTextUNet, a multimodal segmentation framework that integrates CLIP-based textual embeddings into a Swin Transformer U-Net. Experiments on the QaTa-COV19 dataset showed consistent improvements over CNN- and Transformer-based baselines, with the four-stage variant offering the best trade-off between accuracy and complexity.  
This study has some limitations: experiments were limited to a single dataset and 2D images, and text annotations may not always be available in clinical settings. In future work, we aim to extend our model to multi-disease and 3D datasets, leverage domain-specific language models, and validate performance in real-world clinical workflows.

\bibliographystyle{IEEEtran}
\bibliography{Reference}
\end{document}